\documentclass[runningheads]{llncs}
\usepackage[T1]{fontenc}
\usepackage{graphicx}
\usepackage{amsfonts}
\usepackage{array}
\usepackage{hyperref}
\usepackage[usenames]{color}

\usepackage{multirow}

\begin{document}
\title{Benchmarking Graph Neural Networks for Document Layout Analysis in Public Affairs}
\titlerunning{Benchmarking Graph Neural Networks for Document Layout Analysis}
\author{Miguel Lopez-Duran\orcidID{0009-0001-0976-7174} \and
Julian Fierrez\orcidID{0000-0002-6343-5656} \and
Aythami Morales\orcidID{0000-0002-7268-4785} \and
Ruben Tolosana\orcidID{0000-0002-9393-3066} \and
Oscar Delgado-Mohatar\orcidID{0000-0001-5124-7597} \and
Alvaro Ortigosa\orcidID{0000-0002-7674-4132}}
\authorrunning{M. Lopez-Duran, J. Fierrez, A. Morales, et al.}
\institute{School of Engineering, Univ. Autónoma de Madrid (UAM), 28049 Madrid, Spain \\
\email{\{miguel.lopezd, julian.fierrez, aythami.morales, ruben.tolosana, oscar.delgado, alvaro.ortigosa\}@uam.es}}
\maketitle              % typeset the header of the contribution
\begin{abstract}
The automatic analysis of document layouts in digital-born PDF documents remains a challenging problem due to the heterogeneous arrangement of textual and nontextual elements and the imprecision of the textual metadata in the Portable Document Format. In this work, we benchmark Graph Neural Network (GNN) architectures for the task of fine-grained layout classification of text blocks from digital native documents. We introduce two graph construction structures: a $k$-closest-neighbor graph and a fully connected graph, and generate node features via pre-trained text and vision models, thus avoiding manual feature engineering. Three experimental frameworks are evaluated: single-modality (text or visual), concatenated multimodal, and dual-branch multimodal. We evaluated four foundational GNN models and compared them with the baseline. Our experiments are specifically conducted on a rich dataset of public affairs documents that includes more than $20$ sources (e.g., regional and national-level official gazettes), $37$K PDF documents, with $441$K pages in total. Our results demonstrate that GraphSAGE operating on the $k$-closest-neighbor graph in a dual-branch configuration achieves the highest per-class and overall accuracy, outperforming the baseline in some sources. These findings confirm the importance of local layout relationships and multimodal fusion exploited through GNNs for the analysis of native digital document layouts.

\keywords{Graph Neural Networks  \and Document Layout Analysis \and Digital document classification.}
\end{abstract}
\section{Introduction}

The increasing volume of digital documents has created a need for accurate, efficient, and scalable algorithms capable of automatically processing the information within the documents. A large proportion of these documents are stored in the Portable Document Format (PDF), a standard originally developed by Adobe and later standardized in 2008~\cite{edition2008document}. The widespread adoption of PDF is largely due to its ability to present documents independently of the software, hardware, or operating system used, as well as its support for encryption, compression, digital signatures, and editing.

As a result of these advantages, PDF has become the standard format for document management in public administrations and private companies. However, the automatic processing of PDF documents remains a challenging task. Extracting and analyzing relevant information requires understanding the structure and relationships between the various elements of the document, especially text blocks, since they provide different information depending on their semantic role within the document.

Document Layout Analysis (DLA) addresses this challenge by aiming to detect and classify the fundamental components of a document. This task is essential for the automatic processing of digital documents, as it constitutes a preliminary step in the extraction of knowledge and enables further processing with other techniques such as Natural Language Processing (NLP) algorithms~\cite{josi2022preparing,brown2020language,kenton2019bert}. 

As mentioned above, understanding the interactions between the different layout components within a document is essential to accurately determine its semantic structure. As demonstrated by numerous studies in various domains such as social networks~\cite{li2023survey}, or transaction networks~\cite{li2022ttagn}, graphs represent a particularly effective tool for modeling these relationships and have been extensively studied in recent years~\cite{leus2023graph}. Among the techniques for processing graph-structured data, Graph Neural Networks (GNNs) are one of the most prominent approaches~\cite{waikhom2023survey}. GNNs have been used for various tasks in document understanding such as text summarization \cite{xie2023knowledge} or document object detection~\cite{banerjee2024graphkd}.

In this work, we explore the use of GNN-based models combined with multiple graph representations. We use official digital native documents from Spanish public administrations to classify the textual components of each document into specific semantic categories such as title or body. Our approach integrates both text and visual information \cite{2023_SNCS_Human_Pena} in order to enhance the classification of each layout element through specialized models. Furthermore, we evaluated the performance of different GNN architectures with various experimental frameworks to assess the contribution of each component of our method and evaluate its effectiveness.

The main contributions of this work are summarized as follows:

\begin{itemize} 
\item We propose a multimodal graph-based methodology for general Document Layout Analysis, and apply it to Spanish digital native official document preprocessing and representation, leveraging pre-trained models to generate node embeddings, and thus avoiding the need for handcrafted features.
\item A benchmark for Document Layout Analysis (DLA) is presented on digitally conceived documents using GNNs. In this benchmark, we explore two different document graph structures and evaluate the performance of four baseline GNN architectures. We evaluated our approach with three different frameworks to assess the need for each component of our models. 

\item Our proposed approaches are compared against a strong baseline in the DLA of public affairs documents from Spanish administrations. The experiments are carried out on a dataset of public affairs documents that includes 20 sources, $37$K PDF documents, with $441$K pages in total.\footnote{\url{https://github.com/BiDAlab/PALdb}} Our models achieve a higher classification accuracy per class for some classes in most document sources and improve the overall accuracy in one third of them.
\end{itemize}

The paper is organized as follows. In Section \ref{sec::related works} we review the state of the art on multimodal DLA and graph-based DLA methods. Section \ref{sec::methodology} provides the details of the document graph construction and the proposed approaches. The dataset used and the results of our experiments are reported in Section \ref{sec::experiments and results}. Finally, the conclusions and future directions of the work are drawn in Section \ref{sec::conclusions and future work}.

\section{Related Works}
\label{sec::related works}

\subsection{Document Layout Analysis}

The literature on DLA typically distinguishes between two categories of PDF documents: (1) \textit{digitally conceived documents}, which are generated directly from digital sources, and (2) \textit{document images}, which are scanned versions of physical or digital documents (possibly including handwritten text and graphics). 

Traditionally, DLA tools targeting digital-born documents have focused primarily on text extraction. These tools are constrained by the structure of the PDF format, which often lacks semantic metadata, making the identification of layout elements such as tables particularly challenging. Bast \textit{et al.}~\cite{bast2017benchmark} introduced a benchmark for evaluating text extraction methods in digital PDFs, compiling a dataset of more than 10K \textit{arXiv} articles annotated by parsing the associated TeX source files. Zhong \textit{et al.}~\cite{zhong2019publaynet} proposed an automated annotation approach for digital PDF articles by aligning the output of the PDFMiner library with the XML representations of scientific publications. All of these approaches focus on the detection and extraction of the layout components, especially tables, but they lack classification of the semantics of the layout components detected.

\subsection{Multimodal Layout Analysis}

Early work on layout-related tasks relied only on textual or visual information, failing to take advantage of the strong complementarity between these two modalities. Recent work has shown the effectiveness of combining visual, textual, and other types of features such as structural information to improve model expressiveness and performance.

For example, Gu \textit{et al.}~\cite{gu2022xylayoutlm} extract textual, visual, and layout features to address document understanding tasks using a multimodal transformer architecture. Similarly, Cheng \textit{et al.}~\cite{da2023vision} proposed a vision-grid transformer that integrates visual and textual inputs, leveraging multimodal fusion 
\cite{2018_INFFUS_Fierrez,2023_SNCS_Human_Pena} and pre-training strategies to improve downstream performance. In~\cite{zhang2024m2doc}, Zhang \textit{et al.} outline the limitations of previous multimodal DLA approaches and propose a dual multimodal fusion strategy for DLA.

\subsection{GNN-based Document Layout Analysis}

GNNs have shown strong performance in a variety of document-related tasks~\cite{wu2023graph}. Their main strength lies in their ability to model and exploit relational information between entities, which is especially useful in structured document environments.

In the context of DLA, GNNs have been successfully applied to tasks such as table detection and extraction~\cite{qasim2019rethinking,gemelli2022graph}, document classification~\cite{zhang2020every}, and multimodal document understanding~\cite{tang2023unifying}.

However, most existing approaches focus on scanned document images or perform classification at the document level. In contrast, our goal is to classify individual text blocks within digital-born PDF documents from Spanish official sources, leveraging both semantic, visual, and structural information. One advantage of using digital-born documents is the fact that the layout objects are exactly placed, so there is no need of Optical Character Recognition (OCR) assessment, which is not true for scanned documents. To the best of our knowledge, this is the first work to apply a GNN-based method for DLA to official Spanish digital-born documents.

\begin{figure*}[]
  \centering
  \begin{tabular}{cc}
    \multicolumn{2}{c}{\includegraphics[width=0.5\textwidth]{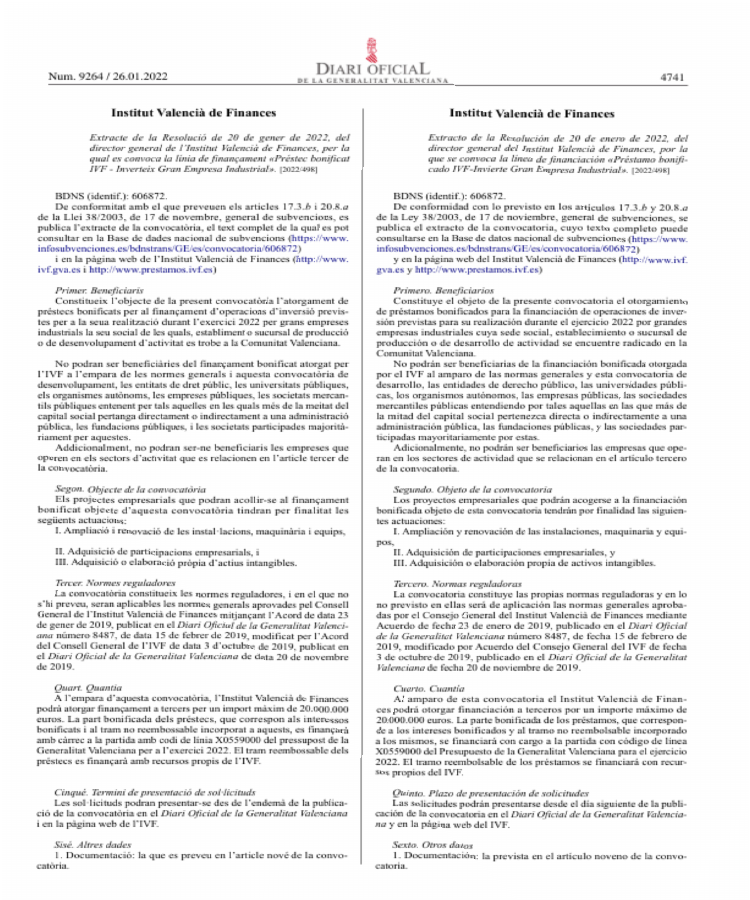}} \\
    \multicolumn{2}{c}{(a)} \\
    \includegraphics[width=0.5\textwidth]{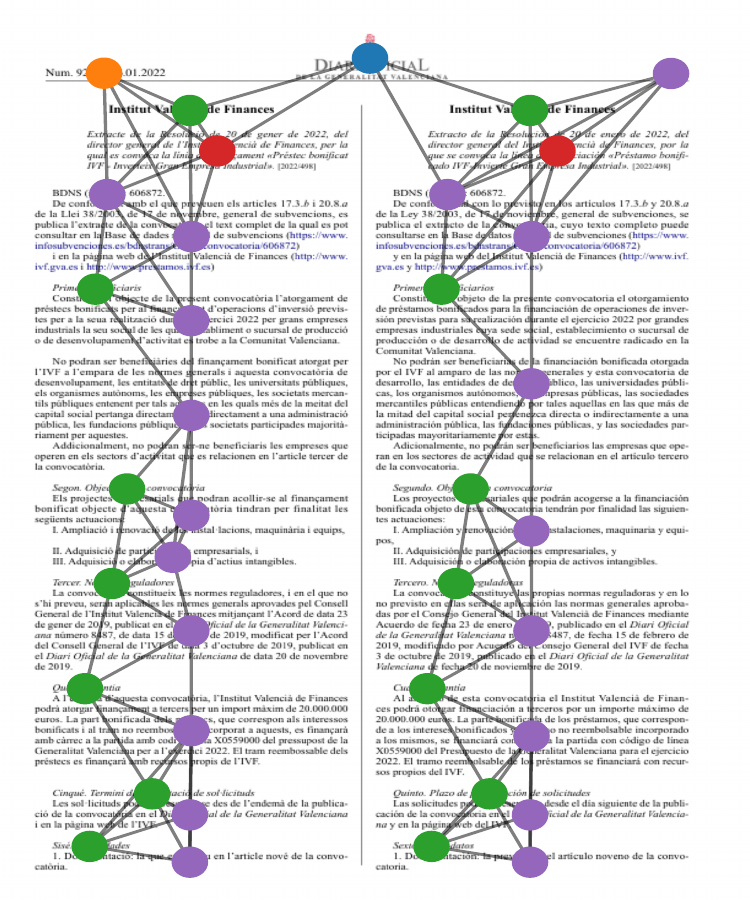} & \includegraphics[width=0.5\textwidth]{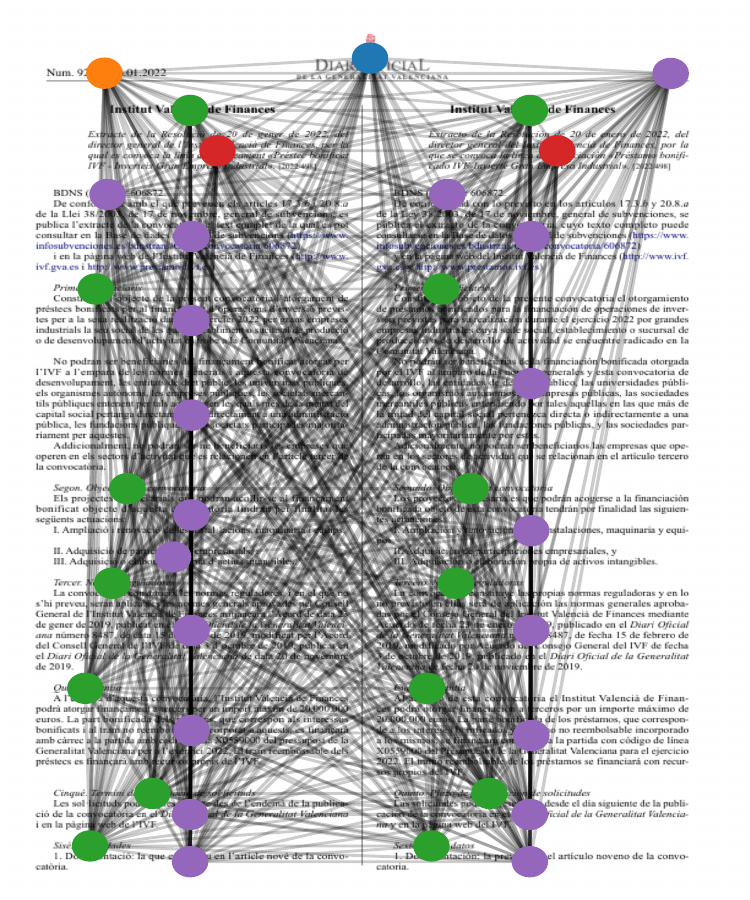}\\
    (b) & (c) 
  \end{tabular}
    
  \caption{Illustration of the graph construction process applied to a document from the \textit{Diario Oficial de la Generalitat Valenciana}. Original document page (a), $k$-closest graph (b), and complete graph (c) representation of the page.}
  \label{example document graph creation}
\end{figure*}
\section{Graph Construction and Proposed Frameworks}
\label{sec::methodology}

In this section, we describe the methodology followed to represent document layouts as graph structures and the approaches adopted to compare various GNN models against the baseline methods.

\subsection{Graph Construction and Initial Features}

A graph $G = (V, E)$ is a mathematical structure consisting of a set of nodes or vertices, denoted $V$, and a set of edges or links, denoted $E$, which define the connections between the nodes.

Given a document $D$ comprising $P$ pages, we represent each page $P_i$ using two distinct graphs: $G_i^K = (V_i, E_i^K)$ and $G_i^C = (V_i, E_i^C)$. In both cases, the set of nodes $V_i$ corresponds to the layout objects within the page.

The construction of edges differs between the two graph structures. For $G_i^K$, which we refer to as the $k$-closest graph, edges are established by connecting each node to its $k$ closest neighbors, where proximity is determined based on the Euclidean distance between the centroids of the layout objects within the page (see Figure~\ref{example document graph creation}(b)). In all of our experiments, we set $k = 4$ in order to focus on local dependencies while keeping the diameter of the graph small, to easily connect distant nodes. This approach is motivated by the hypothesis that the category of a layout object is significantly influenced by its neighboring elements. In contrast, $G_i^C$, which is called the complete graph, is designed as a fully connected graph, where each node is linked to all other nodes within the graph (see Figure~\ref{example document graph creation}(c)). This structure is intended to minimize the introduction of expert-driven biases, allowing the model to generalize effectively across various document layouts.

After constructing the graphs for each page, each document is represented as two sequences of graph structures of the same type, denoted $\{ G_i^j \}_{i = 1}^P$, where $j \in \left\{ K, C \right\}$ and $P$ corresponds to the number of pages in the document.

Once the document graph representation is established, initial node features are generated to use them as input for the models. Rather than relying on the hand-crafted features provided in the dataset we used (see Section \ref{sec::dataset}), which are partially incomplete for images, tables, and URLs, we employ two pretrained models to effectively extract textual and visual information from each layout object.

For text embedding, based on previous work by Peña et al.~\cite{pena2023leveraging}, we employ \textit{RoBERTa-base}~\cite{gutierrez2022maria} as the backbone model. This model is available in the \textit{HuggingFace} repository\footnote{\url{https://huggingface.co/PlanTL-GOB-ES}}. \textit{RoBERTa} models~\cite{liu2019roberta} are variants of \textit{BERT}~\cite{kenton2019bert} that incorporate optimized hyperparameter selection and an improved pretraining strategy. \textit{RoBERTa-base}, specifically, is a Spanish-language version of these models, pretrained on a corpus comprising 570 GB of clean Spanish text. 

Given a layout object, we process its textual information as follows. If the object corresponds to a text block, we apply the RoBERTa-base model directly to its raw textual content and extract the embedding of the [CLS] token as its representation. For images, due to the absence of textual information, we use the string "0" as input to the model. In the case of links, the URL itself is passed as a string. For tables, we extract their contents, flatten them into a single-line text representation, process the resulting text through the transformer model, and use the final classification token ([CLS]) embedding as the textual table representation. After applying this process, we obtain the text embedding of each layout object, denoted as $\mathbf{x}_{T}$.

For visual embedding, we employ a \textit{ResNet} model~\cite{he2016deep} as the backbone. Residual Networks (\textit{ResNet}) leverage residual learning and identity mappings with shortcut connections to effectively capture representative features from images. To ensure efficient and lightweight preprocessing, we select ResNet-18.

Given a layout object and its corresponding PDF document page, we extract the visual representation by cropping the PDF page according to the bounding box of the object. We then apply the vision model excluding the final classification head to the cropped image. The resulting visual embedding is denoted as $\mathbf{x}_{V}$.

\subsection{Proposed Frameworks}

To assess the performance of various GNN models, we propose three distinct frameworks in which GNNs serve as the backbone. Although images, tables, and links are not used directly during training, their embeddings are updated and retained within the graphs. This decision is based on the hypothesis that these elements contribute structural information relevant to the semantic interpretation of their neighboring text blocks.

All proposed frameworks are designed for a supervised node classification task, where each text block is categorized into one of four classes: Identifier, Summary, Title, or Body. The models consist of one or two branches in which node features are processed through a GNN backbone model. To evaluate the performance and suitability of these frameworks, and determine whether specialized GNN models can be developed for this task, we select four fundamental GNN architectures:

\begin{itemize}
\item \textbf{Graph Convolutional Network (GCN)}~\cite{kipf2016semi}: One of the first GNN architectures, designed to learn representative node and graph features in a semi-supervised manner.
\item \textbf{Graph Attention Network (GAT)}~\cite{velivckovic2018graph}: One of the most widely used graph architectures employs an attention mechanism to learn the relative importance of neighboring nodes for the downstream task.
\item \textbf{GraphSAGE}~\cite{hamilton2017inductive}: An inductive architecture that updates the node embeddings by efficiently aggregating features from the local neighborhood of each node.
\item \textbf{Topology Adaptive Graph Convolutional Network (TAGCN)}~\cite{du2017topology}: A GNN variant that adaptively processes graphs to learn across different topologies and effectively captures long-range dependencies.
\end{itemize}

The three proposed frameworks in which we will test the four previous GNNs differ in the type of features utilized (textual and vision) and in the number of processing branches. The defined frameworks are as follows: a single branch using only one type of feature, a single branch incorporating both textual and vision features, and a dual-branch architecture leveraging both feature types. The specific classification strategy varies depending on the framework.

\subsubsection{Single Feature Type, Single Branch.}

In this framework, each document page node is processed through the GNN model using only one type of feature (textual or visual), while discarding the other. A fully connected layer is then applied to classify each layout object on the page, represented as: 
\begin{equation}
    \label{label framework 1}
    \mathbb{P}_1(\omega | \mathbf{x}_{f}),\, f \in \left\{ T, V\right\}.
\end{equation}

\subsubsection{Both Feature Types, Single Branch.}

In this framework, the initial embedding of the initial features of each document page node is the concatenation of both the text and the vision features, denoted as $\mathbf{x} = (\mathbf{x}_{T} \parallel \mathbf{x}_{V})$, which is passed through the GNN model, allowing the model to learn both representations simultaneously. A fully connected layer is applied to classify each layout object on the page, resulting in:
\begin{equation}
    \label{label framework 2}
    \mathbb{P}_2(\omega | \mathbf{x}).
\end{equation}

\subsubsection{Both Feature Types, Dual Branch.}

In this framework, each feature type is processed independently through dedicated models, creating specialized textual and visual representations. The outputs of both models are then concatenated and passed through two fully connected layers to classify each layout object on the page, which results in:
\begin{equation}
    \label{label framework 3}
    \mathbb{P}_3(\omega | \mathbf{x}_{T}, \mathbf{x}_{V}),
\end{equation}
where $\omega \in \left\{ \textit{ID}, \textit{Title}, \textit{Summary}, \textit{Body} \right\}$ in all settings.

\begin{figure}
\centering
\includegraphics[scale=0.35, angle=-90]{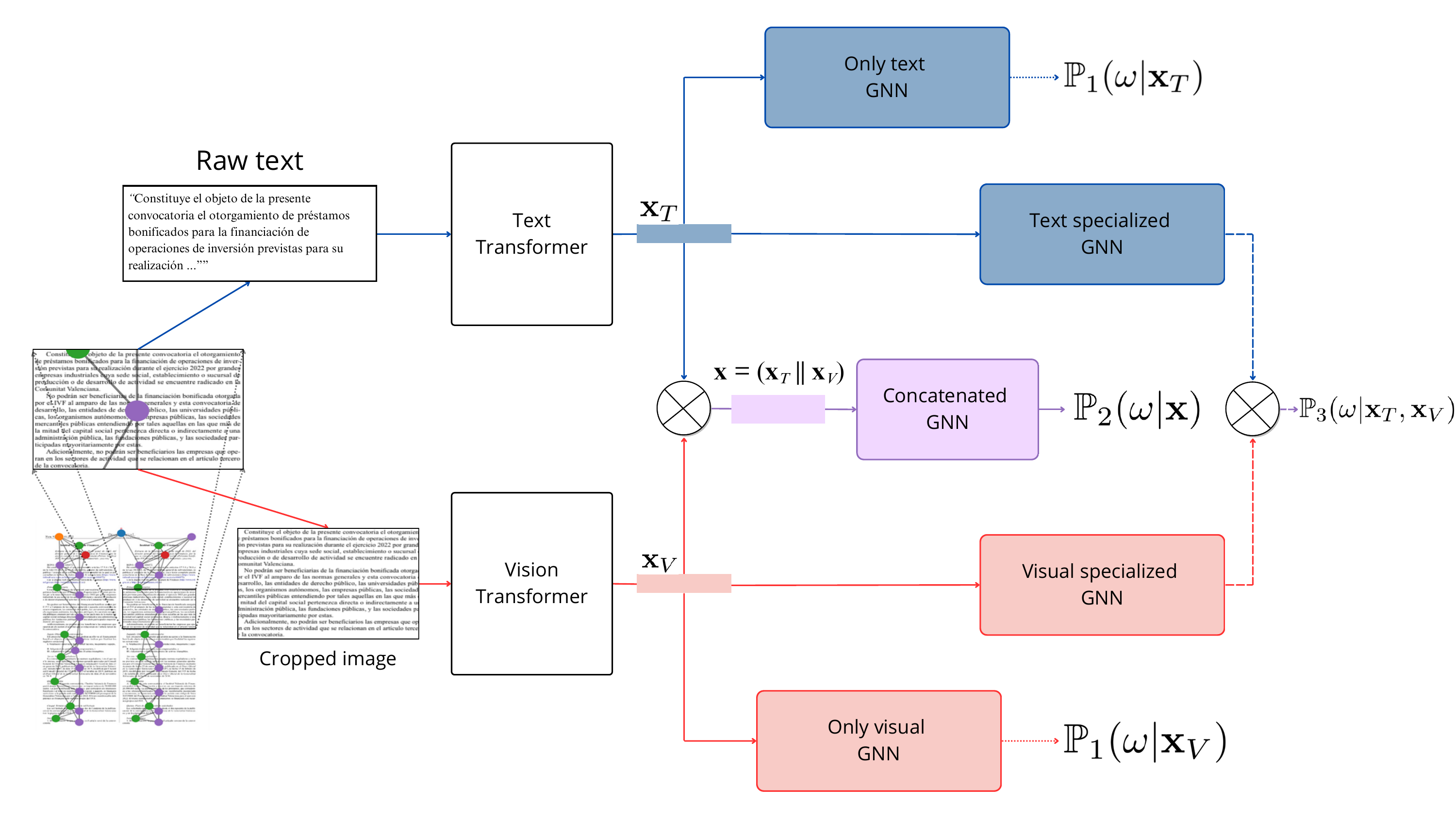}
\caption{Overview of the three experimental frameworks. After passing each object through the transformers, four different models are trained independently to perform three different classifications of the text block object.} 
\label{fig2}
\end{figure}

An overview of the different frameworks is shown in Figure~\ref{fig2}.

In all frameworks, we perform a train/test split at the document level and execute the forward pass at the page level to mitigate intra-document biases, as observed in \cite{pena2024continuous}. The selected GNN architectures are applied in a sequence of layers, each followed by batch normalization, an Exponential Linear Unit (ELU) activation function, and a dropout layer to enhance performance.

\section{Experiments and Results}
\label{sec::experiments and results}
In this section, we describe the dataset we used to evaluate the models. The experimental setup and the results of our approaches are also presented. The primary objective of our experiments is to assess the effectiveness of multimodality and the usefulness of GNNs in the layout classification task for official documents.

\subsection{Public Affairs Layout (PAL) Dataset}
\label{sec::dataset}

The Public Affairs Layout (PAL) dataset~\cite{pena2024continuous} contains over 37.9K PDF document layouts extracted from a total of 24 official Spanish gazettes.\footnote{\url{https://github.com/BiDAlab/PALdb}} The documents were annotated using a human-in-the-loop approach, assigning each layout object one of the following labels: Image, Table, Link, Identifier, Title, Summary, or Body. Although the first three categories can be directly retrieved from the internal PDF metadata, the remaining four are initially classified as generic text blocks. This limitation highlights the need to develop algorithms and models capable of accurately distinguishing semantic differences between these text-based categories.

\begin{figure}[t!]
    \centering
    \includegraphics[width = 0.98\textwidth]{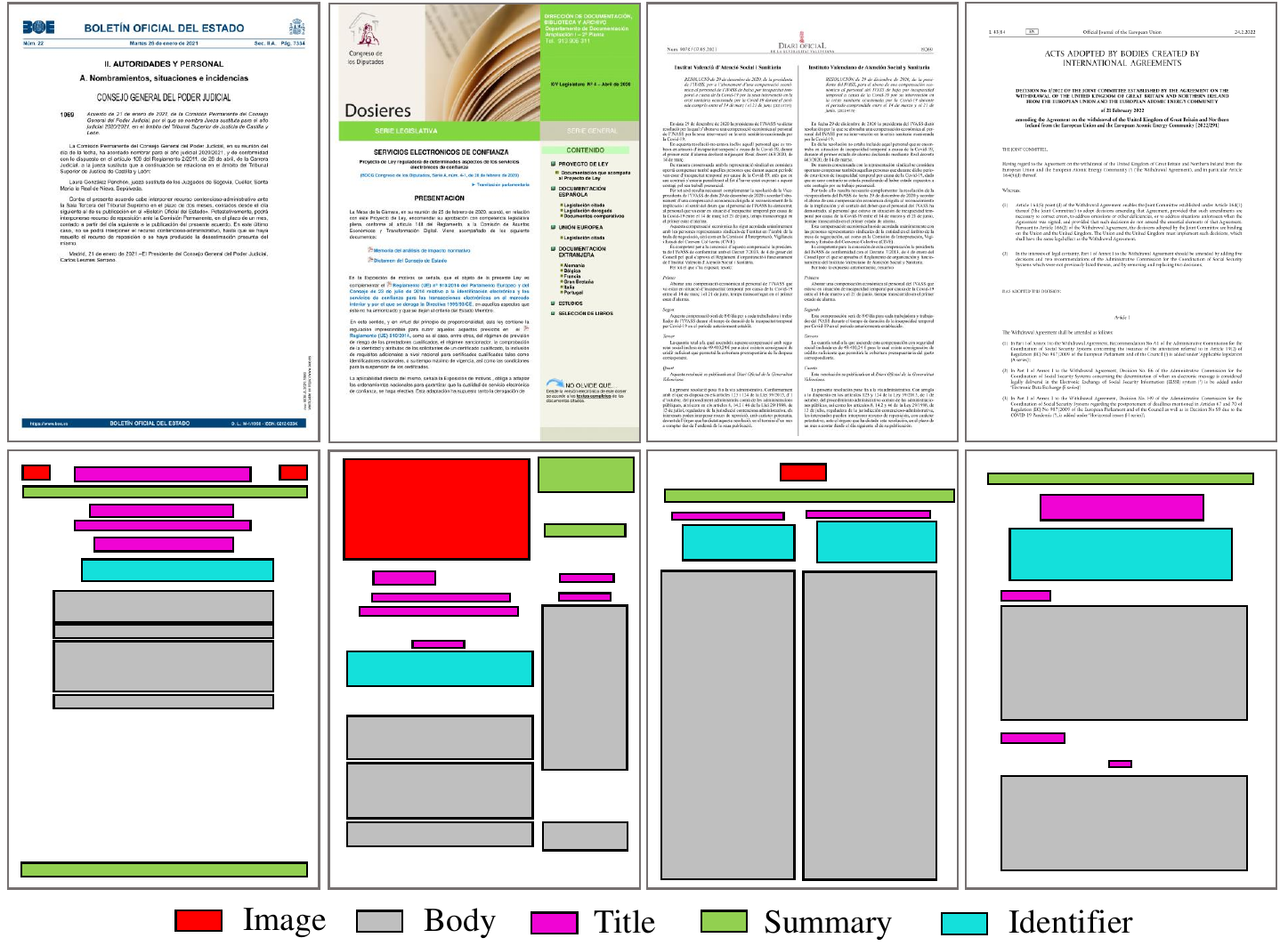}
    \caption{Visual examples of page document images from different official gazettes (upper row). From left to right we present: i) Spanish official gazette (i.e., BOE); ii) public affairs document with the results of a vote in the Spanish Parliament; iii) Spanish regional official gazette (i.e. DOGV); and iv) Official Journal of the European Union. The bottom row presents illustrations of their layout components simplified by colored block (i.e., each color represents a document layout component).}
    \label{fig:layout_examples}
\end{figure}

The PAL database is divided into two subsets: a training set, originally used to train the Random Forest models introduced in~\cite{pena2024continuous} to classify the text blocks, and a validation set, initially used for model evaluation and subsequently verified by a human annotator. In all of our experiments, we exclusively use the validation set. This decision is motivated by the fact that the training set may contain labeling inaccuracies, as it was not externally validated. For detailed statistics on the validation set, we refer to the original publication. Some examples of documents and layouts from the dataset are presented in Figure~\ref{fig:layout_examples}.

\subsection{Experimental Protocol}
In all experiments, we train a separate model for each source, where the source corresponds to one of the official gazettes included in the PAL database. Following the protocol established by Peña \textit{et al.}~\cite{pena2024continuous}, we use 80\% of the documents in the validation set from each source for training and the remaining 20\% for testing. We employ a $k$-fold cross-validation strategy with $k = 5$ and random initializations. Given the significant imbalance in the number of documents across sources, the batch size is manually adjusted per source to ensure stable and consistent training.

All models were trained for 350 epochs with a weighted cross-entropy loss, and the stochastic gradient descent (SGD) optimizer with a learning rate of $10^{-3}$ and a momentum of $0.9$ and a dropout of $0.1$.

\subsection{Results}
\begin{table}[t]
\caption{Overall and per-class mean classification accuracy across all folds and the 24 sources compared against the baseline. We report the best performing models for Only Text (OT) and Only Visual (OV) features, and Concatenated (Conc.) framework. We report the six best performing models for dual setting, where the first model refers to the text specialization and the second model to the visual specialization. G = GCN, S = GraphSAGE, T = TAGCN, RF = Random Forest.}
\label{tab1}
\centering
\setlength{\tabcolsep}{2pt} 
\renewcommand{\arraystretch}{1.2} 
\begin{tabular}{
    >{\raggedright\arraybackslash}p{1.0cm} 
    >{\centering\arraybackslash}p{1.8cm} 
    >{\centering\arraybackslash}p{1.8cm} 
    >{\centering\arraybackslash}p{1.3cm} 
    >{\centering\arraybackslash}p{1.3cm} 
    >{\centering\arraybackslash}p{1.3cm} 
    >{\centering\arraybackslash}p{1.3cm} 
    >{\centering\arraybackslash}p{1.3cm} }
\hline 

Model                 & Backbone  & Type of graph & \multicolumn{5}{c}{Classification   accuracy (\%)}  \\ \cline{4-8} 
                      &           &               & Overall   & ID      & Title   & Summary   & Body   \\ \hline 
\cite{pena2024continuous}                  & RF        & -             & $97.48$     & $99.81$   & $93.70$   & $95.81$     & $98.25$  \\ \hline 
OT                    & GraphSAGE & $k$-closest     & $95.44$     & $99.01$   & $93.20$    & $92.13$     & $95.52$  \\ 
OV                    & GraphSAGE & $k$-closest      & $75.97$     & $81.76$   & $65.03$   & $41.47$     & $81.29$  \\ \hline 
Conc.          & GraphSAGE & Complete      & $95.03$     & $98.15$   & $92.12$   & $81.99$     & $96.09$  \\ \hline 
\multirow{5}{*}{Dual} & S+S       & $k$-closest     & $95.85$     & $98.86$   & $92.42$   & $89.97$     & $96.42$  \\ 
                      & S+S       & Complete      & $95.98$     & $99.32$   & $93.36$   & $82.63$     & $96.92$  \\ 
                      & S+G       & $k$-closest     & $95.81$     & $98.91$   & $92.88$   & $91.03$     & $96.29$  \\ 
                      & S+T       & $k$-closest     & $95.10$     & $98.56$   & $91.16$   & $90.15$     & $96.03$  \\ 
                      & S+T       & Complete      & $95.84$     & $99.27$   & $93.04$   & $83.06$     & $96.56$  \\ \hline
\end{tabular}
\end{table}

The general results for the proposed frameworks are presented in Table~\ref{tab1}, which summarizes the graph structures and models that perform the best, together with a comparison to the baseline. Given the large number of document sources and experimental settings, we report the average classification accuracy across all folds and sources. For the first two frameworks (Only Text or Only Visual, and Concatenated Features), we report the best-performing model, while for the dual-branch framework, we report the top five model combinations, specifying the graph type and GNN backbones used.

Our analysis reveals that the $k$-closest graph structure consistently yields better results, outperforming the complete graph structure in five of the eight top-performing configurations. Additionally, GraphSAGE emerges as the most effective backbone, present in all eight configurations, either for text or visual features in the dual-branch setting. These findings support our initial hypothesis regarding the importance of local layout relationships in determining the semantic category of each text block. Furthermore, the dual-branch framework outperforms the others in four of the five top configurations, confirming the benefit of processing text and visual features separately before classification.

However, these results may present an overly optimistic view, as they are based on selecting the best configuration for each framework independently and reporting averaged scores across all folds and sources. To provide a more detailed and realistic evaluation, we assess the performance of a fixed configuration in five representative sources. Based on the previous results, we selected the $k$-closest graph structure and the dual-branch framework with GraphSAGE, as it was the configuration that performed the best on average. The selected sources correspond to those with the largest number of documents in the validation set. The results presented in Table~\ref{tab2} highlight the fields in which our models outperform the baseline method proposed by Peña \textit{et al.}~\cite{pena2024continuous}. 

\begin{table}[t]
\caption{Overall and per-class classification accuracy for $k$-closest graph and the best model selection containing GraphSAGE for the five selected sources. We report the accuracy in terms of $mean_{std}$ across five folds of cross validation. The first model refers to the text specialization and the second model to the visual specialization. S = GraphSAGE, T = TAGCN. The \href{www.boe.es}{BOE} (Boletín Oficial del Estado) source is the Official State Bulletin where the Spanish government publishes laws, regulations, and official acts that must be made public. The remaining sources correspond to the official gazettes of other regions within the Spanish state. \href{https://sede.asturias.es/servicios-del-bopa}{BOPA} = Boletín Oficial del Principado de Asturias, \href{https://www.euskadi.eus/web01-bopv/es/}{BOPV} = Boletín Oficial del Páis Vasco, \href{https://bomemelilla.es/bomes/2025}{BOME} = Boletín Oficial de Melilla, \href{https://w123.bcn.cat/APPS/egaseta/home.do?reqCode=init}{BOAB} = Boletín Oficial del Ayuntamiento de Barcelona.}
\label{tab2}
\centering
\setlength{\tabcolsep}{2pt} 
\renewcommand{\arraystretch}{1.2} % Aumenta altura de filas
\begin{tabular}{>{\raggedright\arraybackslash}p{1.0cm} 
    >{\centering\arraybackslash}p{1.8cm} 
    >{\centering\arraybackslash}p{1.5cm} 
    >{\centering\arraybackslash}p{1.5cm} 
    >{\centering\arraybackslash}p{1.5cm} 
    >{\centering\arraybackslash}p{1.5cm}
    >{\centering\arraybackslash}p{1.5cm}}
\hline
Source & Backbone & \multicolumn{5}{c}{Classification   accuracy (\%)} \\ \cline{3-7}
                        &                                  & Overall         & ID          & Title       & Summary     & Body        \\ \hline
BOE                     & S+S                              & $97.93_{2.03}$ & $99,52_{0,76}$ & {\boldmath $97,77_{1,84}$} & $98,45_{1,27}$ & $97,45_{3,56}$ \\
BOPA                    & S+T                              & $95,91_{0,92}$     & $100_{0}$      & $95,50_{1,81}$ & $93,43_{7,66}$ & $95,73_{1,26}$ \\
BOPV                    & S+S                              & {\boldmath$98,40_{0,81}$}     & $99,72_{0,56}$ & {\boldmath$98,59_{1,75}$} & {\boldmath$99,09_{1,82}$} & $97,85_{0,85}$ \\
BOME                    & S+T                              & $96,08_{0,87}$     & $100_{0}$      & $92,54_{4,80}$ & $90,51_{8,63}$ & $97,07_{1,81}$ \\
BOAB                    & S+S                              & $97,83_{1,59}$     & $100_{0}$      & {\boldmath$95,83_{2,76}$} & $98,92_{0,89}$ & $98,01_{1,60}$ \\ \hline
\end{tabular}
\end{table}

We observe that the selected model and graph structure consistently perform well across the evaluated sources, particularly with the use of GraphSAGE as the backbone for both textual and vision models. The results demonstrate strong performance in all five selected sources, with our approach exceeding the baseline in the classification of specific classes in four of the selected sources and achieving a higher overall accuracy in one of the cases.

\section{Conclusions and Future Work}
\label{sec::conclusions and future work}

In this work, we introduced several experimental frameworks to evaluate the performance of different graph-based models and graph structures for the semantic classification of text blocks in digital-born PDF documents from the public affairs domain.

We represent each document page as a graph, where nodes correspond to layout objects and edges are defined using two alternative structures: a $k$-closest neighbor graph, where each node is connected to its $k$ closest neighbors on the page, and a complete graph, where all nodes are connected to each other. Thus, each document is modeled as a sequence of graphs, and classification is performed at the text block level. To avoid manual feature engineering, we initialize node embeddings using pretrained models for both text and vision features.

We proposed and evaluated three experimental frameworks: one that uses text or vision features independently, a second that concatenates both types of feature in a single model, and a third that trains separate models for each modality and subsequently fuses their outputs for classification. Our experiments in the PAL dataset~\cite{pena2024continuous} demonstrate strong performance, particularly for the dual-branch framework with the GraphSAGE architecture, which outperforms the baseline in per-class accuracy for several sources and overall accuracy in one third of them, while maintaining consistency between different subsets of the data.

For future work, several research directions remain open. These include enhancing the representation of text and vision features, incorporating additional information such as relative position within the page (an approach that has proven effective in scanned document settings~\cite{gemelli2022doc2graph}) and exploring more advanced graph models to capture deeper structural relationships. Furthermore, the application of sequential Graph Neural Networks, which have already been used for spatiotemporal forecasting~\cite{liang2024dynamic}, or alternative graph structures, such as visibility graphs~\cite{riba2022table}, could further improve the semantic classification of layout elements in digital-born PDF documents. Furthermore, we also plan to explore DLA in combination with Named Entity Recognition (NER) and Entity Linking (EL) technologies to enhance the information extracted from the document's text, a proposal that has already been applied to privacy-preserving text classification tasks~\cite{kutbi2023named,mancera2025pba} and form understanding~\cite{dang2021end}. Finally, addressing biases \cite{2023_ECAIw_LFIT-XAI_Tello,peña2025bias} and facilitating audit tools \cite{mancera2025my} in this research line on DLA/NER/EL is also part of our research agenda.

\begin{credits}
\subsubsection{\ackname}
Supported by HumanCAIC (TED2021-131787B-I00 MICINN), BBforTAI (PID2021-127641OB-I00 MICINN/FEDER), M2RAI (PID2024-160053OB-I00 MICIU/FEDER), Cátedra ENIA UAM-VERIDAS en IA Responsable (NextGenerationEU PRTR TSI-100927-2023-2), and Research Agreement 
DGGC/UAM/FUAM for Biometrics and Applied AI. Morales is also supported by the Madrid Government in the line of Excellence for University Teaching Staff (V PRICIT). Work conducted within the ELLIS Unit Madrid.
\subsubsection{\discintname} The authors declare no conflict of interests.
\end{credits}

\bibliographystyle{splncs04}
\bibliography{main}

\begin{thebibliography}{10}
\providecommand{\url}[1]{\texttt{#1}}
\providecommand{\urlprefix}{URL }
\providecommand{\doi}[1]{https://doi.org/#1}

\bibitem{banerjee2024graphkd}
Banerjee, A., Biswas, S., Llad{\'o}s, J., Pal, U.: {GraphKD: Exploring Knowledge Distillation Towards Document Object Detection with Structured Graph Creation}. In: Proc. of ICDAR. pp. 354--373. Springer (2024)

\bibitem{bast2017benchmark}
Bast, H., Korzen, C.: {A benchmark and evaluation for text extraction from PDF}. In: ACM/IEEE Joint Conf. on Digital Libraries (JCDL). pp. 1--10 (2017)

\bibitem{brown2020language}
Brown, T., Mann, B., Ryder, N., Subbiah, M., Kaplan, J.D., Dhariwal, P., Neelakantan, A., Shyam, P., Sastry, G., Askell, A., et~al.: {Language models are few-shot learners}. Advances in Neural Information Processing Systems  \textbf{33},  1877--1901 (2020)

\bibitem{da2023vision}
Da, C., Luo, C., Zheng, Q., Yao, C.: {Vision grid transformer for document layout analysis}. In: Proc. ICCV. pp. 19462--19472 (2023)

\bibitem{dang2021end}
Dang, T.A.N., Hoang, D.T., Tran, Q.B., Pan, C.W., Nguyen, T.D.: End-to-end hierarchical relation extraction for generic form understanding. In: 2020 25th International Conference on Pattern Recognition (ICPR). pp. 5238--5245. IEEE (2021)

\bibitem{du2017topology}
Du, J., Zhang, S., Wu, G., Moura, J.M., Kar, S.: {Topology adaptive graph convolutional networks}. arXiv preprint arXiv:1710.10370  (2017)

\bibitem{2018_INFFUS_Fierrez}
Fierrez, J., Morales, A., Vera-Rodriguez, R., Camacho, D.: Multiple classifiers in biometrics. {Part} 2: Trends and challenges. Information Fusion  \textbf{44},  103--112 (2018)

\bibitem{gemelli2022doc2graph}
Gemelli, A., Biswas, S., Civitelli, E., Llad{\'o}s, J., Marinai, S.: {Doc2graph: a task agnostic document understanding framework based on graph neural networks}. In: European Conference on Computer Vision. pp. 329--344. Springer (2022)

\bibitem{gemelli2022graph}
Gemelli, A., Vivoli, E., Marinai, S.: {Graph neural networks and representation embedding for table extraction in PDF documents}. In: 2022 26th International Conference on Pattern Recognition (ICPR). pp. 1719--1726. IEEE (2022)

\bibitem{gu2022xylayoutlm}
Gu, Z., Meng, C., Wang, K., Lan, J., Wang, W., Gu, M., Zhang, L.: {XYLayoutLM: Towards layout-aware multimodal networks for visually-rich document understanding}. In: Proceedings of the IEEE/CVF Conference on Computer Vision and Pattern Recognition. pp. 4583--4592 (2022)

\bibitem{gutierrez2022maria}
Guti{\'e}rrez~Fandi{\~n}o, A., et~al.: {MarIA: Spanish Language Models}. Procesamiento del Lenguaje Natural  \textbf{68} (2022)

\bibitem{hamilton2017inductive}
Hamilton, W., Ying, Z., Leskovec, J.: {Inductive representation learning on large graphs}. Advances in Neural Information Processing Systems  \textbf{30} (2017)

\bibitem{he2016deep}
He, K., Zhang, X., Ren, S., Sun, J.: {Deep residual learning for image recognition}. In: Proc. CVPR. pp. 770--778 (2016)

\bibitem{edition2008document}
ISO: {Document management—-Portable document format—-Part 1: PDF 1.7. Standard}  (2008)

\bibitem{josi2022preparing}
Josi, F., Wartena, C., Heid, U.: {Preparing legal documents for NLP analysis: Improving the classification of text elements by using page features}. In: Computer Science \& Information Technology (CS \& IT). pp. 17--29. AIRCC Publishing Corporation (2022)

\bibitem{kenton2019bert}
Kenton, J.D.M.W.C., Toutanova, L.K.: {BERT: Pre-training of deep bidirectional transformers for language understanding}. In: Proceedings of NAACL-HLT. vol.~1. Minneapolis, Minnesota (2019)

\bibitem{kipf2016semi}
Kipf, T.N., Welling, M.: {Semi-supervised classification with graph convolutional networks}. arXiv preprint arXiv:1609.02907  (2016)

\bibitem{kutbi2023named}
Kutbi, M.: Named entity recognition utilized to enhance text classification while preserving privacy. IEEE Access  \textbf{11},  117576--117581 (2023)

\bibitem{leus2023graph}
Leus, G., Marques, A.G., Moura, J.M., Ortega, A., Shuman, D.I.: {Graph signal processing: History, development, impact, and outlook}. IEEE Signal Processing Magazine  \textbf{40}(4),  49--60 (2023)

\bibitem{li2022ttagn}
Li, S., Gou, G., Liu, C., Hou, C., Li, Z., Xiong, G.: {TTAGN: Temporal transaction aggregation graph network for ethereum phishing scams detection}. In: Proceedings of the ACM Web Conference 2022. pp. 661--669 (2022)

\bibitem{li2023survey}
Li, X., Sun, L., Ling, M., Peng, Y.: {A survey of graph neural network based recommendation in social networks}. Neurocomputing  \textbf{549},  126441 (2023)

\bibitem{liang2024dynamic}
Liang, G., Tiwari, P., et~al.: {Dynamic Causal Explanation Based Diffusion-Variational Graph Neural Network for Spatiotemporal Forecasting}. IEEE Trans. on Neural Networks and Learning Systems  (2024)

\bibitem{liu2019roberta}
Liu, Y., Ott, M., Goyal, N., Du, J., Joshi, M., Chen, D., Levy, O., Lewis, M., Zettlemoyer, L., Stoyanov, V.: {RoBERTa: A robustly optimized bert pretraining approach}. arXiv preprint arXiv:1907.11692  (2019)

\bibitem{mancera2025my}
Mancera, G., DeAlcala, D., Fierrez, J., Tolosana, R., Morales, A.: Is my text in your {AI} model? {G}radient-based membership inference test applied to {LLMs}. arXiv preprint arXiv:2503.07384  (2025)

\bibitem{mancera2025pba}
Mancera, G., Morales, A., Fierrez, J., Tolosana, R., Penna, A., Lopez-Duran, M., Jurado, F., Ortigosa, A.: {PBa-LLM}: Privacy- and bias-aware {NLP} using named-entity recognition ({NER}). In: IAPR Intl. Conf. on Document Analysis and Recognition Workshops (ICDARw) (September 2025)

\bibitem{pena2024continuous}
Pe{\~n}a, A., Morales, A., Fierrez, J., Ortega-Garcia, J., Puente, I., Cordova, J., Cordova, G.: {Continuous document layout analysis: Human-in-the-loop AI-based data curation, database, and evaluation in the domain of public affairs}. Information Fusion  \textbf{108},  102398 (2024)

\bibitem{pena2023leveraging}
Pe{\~n}a, A., Morales, A., Fierrez, J., Serna, I., Ortega-Garcia, J., Puente, I., Cordova, J., Cordova, G.: {Leveraging large language models for topic classification in the domain of public affairs}. In: International Conference on Document Analysis and Recognition. pp. 20--33. Springer (2023)

\bibitem{peña2025bias}
Peña, A., Fierrez, J., Morales, A., Mancera, G., Lopez, M., Tolosana, R.: Addressing bias in {LLMs}: Strategies and application to fair {AI}-based recruitment. In: AAAI/ACM Conf. on AI, Ethics, and Society (AIES) (October 2025)

\bibitem{2023_SNCS_Human_Pena}
Peña, A., Serna, I., Morales, A., Fierrez, J., Ortega, A., Herrarte, A., Alcantara, M., Ortega-Garcia, J.: Human-centric multimodal machine learning: Recent advances and testbed on {AI}-based recruitment. SN Computer Science  \textbf{4}(5), ~434 (June 2023)

\bibitem{qasim2019rethinking}
Qasim, S.R., Mahmood, H., Shafait, F.: {Rethinking table recognition using graph neural networks}. In: 2019 International Conference on Document Analysis and Recognition (ICDAR). pp. 142--147. IEEE (2019)

\bibitem{riba2022table}
Riba, P., Goldmann, L., Terrades, O.R., Rusticus, D., Forn{\'e}s, A., Llad{\'o}s, J.: {Table detection in business document images by message passing networks}. Pattern Recognition  \textbf{127},  108641 (2022)

\bibitem{tang2023unifying}
Tang, Z., Yang, Z., Wang, G., Fang, Y., Liu, Y., Zhu, C., Zeng, M., Zhang, C., Bansal, M.: {Unifying vision, text, and layout for universal document processing}. In: Proceedings of the IEEE/CVF Conference on Computer Vision and Pattern Recognition. pp. 19254--19264 (2023)

\bibitem{2023_ECAIw_LFIT-XAI_Tello}
Tello, J., de~la Cruz, M., Ribeiro, T., Fierrez, J., Morales, A., Tolosana, R., Alonso, C.L., Ortega, A.: Symbolic {AI (LFIT) for XAI} to handle biases. In: European Conf. on AI Workshops (ECAIw). CEUR-WS, vol.~3523 (2023)

\bibitem{velivckovic2018graph}
Veli{\v{c}}kovi{\'c}, P., Cucurull, G., Casanova, A., Romero, A., Li{\`o}, P., Bengio, Y.: {Graph Attention Networks}. In: ICLR (2018)

\bibitem{waikhom2023survey}
Waikhom, L., Patgiri, R.: {A survey of graph neural networks in various learning paradigms: methods, applications, and challenges}. Artificial Intelligence Review  \textbf{56}(7),  6295--6364 (2023)

\bibitem{wu2023graph}
Wu, L., Chen, Y., Shen, K., Guo, X., Gao, H., Li, S., Pei, J., Long, B., et~al.: {Graph neural networks for natural language processing: A survey}. Foundations and Trends{\textregistered} in Machine Learning  \textbf{16}(2),  119--328 (2023)

\bibitem{xie2023knowledge}
Xie, Q., Tiwari, P., Ananiadou, S.: {Knowledge-enhanced graph topic transformer for explainable biomedical text summarization}. IEEE Journal of Biomedical and Health Informatics  \textbf{28}(4),  1836--1847 (2023)

\bibitem{zhang2024m2doc}
Zhang, N., Cheng, H., Chen, J., Jiang, Z., Huang, J., Xue, Y., Jin, L.: {M2Doc: a multi-modal fusion approach for document layout analysis}. In: Proceedings of the AAAI Conference on Artificial Intelligence. vol.~38, pp. 7233--7241 (2024)

\bibitem{zhang2020every}
Zhang, Y., Yu, X., Cui, Z., Wu, S., Wen, Z., Wang, L.: {Every Document Owns Its Structure: Inductive Text Classification via Graph Neural Networks}. In: Proc. ACL. pp. 334--339 (2020)

\bibitem{zhong2019publaynet}
Zhong, X., Tang, J., Yepes, A.J.: {PubLayNet: largest dataset ever for document layout analysis}. In: 2019 International Conference on Document Analysis and Recognition (ICDAR). pp. 1015--1022. IEEE (2019)

\end{thebibliography}
\end{document}